\documentclass[10pt, letterpaper]{article}

\usepackage[english]{babel}

\usepackage[letterpaper,top=2cm,bottom=2cm,left=3cm,right=3cm,marginparwidth=1.75cm]{geometry}

\usepackage{amsmath, amssymb, amsfonts}
\usepackage{graphicx}
\usepackage[colorlinks=true, allcolors=blue]{hyperref}
\usepackage{booktabs}
\usepackage{cite}

\newcommand{\SE}{SE(3)}
\newcommand{\SO}{SO(3)}
\newcommand{\se}{\mathfrak{se}(3)}

\title{\textbf{SE3Kit: A Lightweight Python Library for Specialized Geometric Primitives in Robotics}}
\author{Daniyal Maroufi$^1$, Omid Rezayof$^1$, and Farshid Alambeigi$^1$}

\date{}

\begin{document}
\maketitle
\thanks{$^1$ $^{1}$D.~Maroufi, O.~Rezayof and F.~Alambeigi are with the Walker Department of Mechanical Engineering and Texas Robotics at The University of Texas at Austin, Austin, TX, 78712, USA. Email: maroufi@utexas.edu, omid.rezayof@utexas.edu, farshid.alambeigi@austin.utexas.edu}

\begin{abstract}
The Python robotics ecosystem faces a challenge: while many libraries exist for rigid body transformations, few are both lightweight and mathematically strict. This paper introduces SE3Kit, a lightweight Python library efficient operations on the Special Euclidean Group $SE(3)$ and the Special Orthogonal Group $SO(3)$. Unlike established frameworks that require heavy dependencies (e.g., SpatialMath, PyPose) or general tools that lack robotics-specific features (e.g., SciPy), SE3Kit targets the gap between these extremes. It is designed for embedded deployment, rapid prototyping, and education while providing rigorous mathematical implementation. It provides a pure-Python, NumPy-only implementation of Lie Group operations, without the overhead of deep learning or other visualization software.
\end{abstract}

Unlike established frameworks such as SpatialMath or {PyPose}, which necessitate extensive dependency chains, SE3Kit utilizes a {NumPy}-only implementation to minimize the computational footprint. This architecture facilitates deployment on resource-constrained embedded systems and streamlines rapid prototyping without the overhead of deep learning or visualization modules. By providing a mathematically rigorous implementation of Lie Group operations in pure Python, \texttt{SE3Kit} offers a scalable solution for both industrial applications and academic research.

\section{Introduction}
The representation of rigid bodies in three-dimensional space is the fundamental mathematical foundation of robotics. Beneath path planners, perception networks, and control loops, roboticists must manipulate elements of the Special Euclidean Group, $\SE$. 

Despite the importance of these concepts, the current software landscape is fragmented. Roboticists are often forced to choose between frameworks that introduce massive dependency chains (e.g., \cite{pypose, spatialmath}), or legacy scripts that lack type safety and active maintenance (e.g., \texttt{transformations.py} \cite{transformationspy}). 

SE3Kit is designed to resolve this fragmentation. It offers a specialized solution: small enough for constrained hardware, yet rigorous enough for advanced robotics research.

\section{State of the Field \& Gap Analysis}
To understand the necessity of SE3Kit, we analyze the limitations of current standard tools in specific contexts.

\subsection{Computational Overhead of Comprehensive Frameworks}
Newer libraries like PyPose \cite{pypose} and Kornia \cite{riba2020kornia} leverage PyTorch to offer differentiability. However, for deterministic robotics tasks (such as parsing sensor streams or computing kinematic chains) this requires substantial computing resources. The overhead of initializing the PyTorch runtime is disproportionate for logic layers that do not require gradient descent, rendering these tools inefficient for standard CPU-based control loops.

\subsection{Computational Overhead of Differentiable Frameworks}
More recent libraries like PyPose \cite{pypose} and Kornia \cite{riba2020kornia} leverage PyTorch to offer differentiability. However, for real-time, deterministic and classical robotics tasks (such as parsing sensor streams or computing kinematic chains) this requires substantial computing resources, making these tools inefficient for standard CPU-based control loops.

\subsection{The Generalist Gap in SciPy}
While \texttt{scipy.spatial.transform} \cite{scipy} has improved with the addition of \texttt{RigidTransform}, it remains a general tool serving broad scientific fields. Yet, it lacks robotics-specific semantics and utilities, specifically:
\begin{itemize}
    \item \textbf{Calibration Solvers:} Essential routines for experimental validation, such as pivot calibration, hand-eye calibration (solving $AX=XB$), and point cloud registration, are domain-specific and absent from general scientific libraries.
    \item \textbf{Lie Algebra Primitives:} Robotics algorithms (e.g., SLAM, EKF) require explicit access to logarithmic ($\log: \SE \to \se$) and exponential maps for optimization. General libraries often hide these mathematical details internally.
\end{itemize}

\section{The SE3Kit Library}
SE3Kit fills the gap for a dependency-minimalist solution, strictly requiring only NumPy.

\subsection{Core Capabilities}
The library implements the algebra of the Special Euclidean Group $\SE$ and the Special Orthogonal Group $\SO$.
\begin{itemize}
    \item \textbf{Unified Representation:} Consistent handling of rotation matrices, quaternions, and Euler angles. This resolves common ambiguities, such as the order of scalar values in quaternions (scalar-first vs. scalar-last).
    \item \textbf{Tangent Space Operations:} Robust implementations of the exponential map and logarithmic map. These are necessary for transforming vector objects like velocities and wrenches between frames.
    \item \textbf{Frame Consistency:} Unlike most libraries that operate on raw arrays, SE3Kit enforces manifold constraints to prevent numerical drift (e.g., ensuring rotation matrices remain orthogonal).
\end{itemize}

\subsection{Architectural Comparison}
Table \ref{tab:comparison} illustrates the ecosystem gap that SE3Kit addresses.

\begin{table}[h]
\centering
\caption{Comparative Analysis of Python Geometric Libraries}
\label{tab:comparison}
\resizebox{\textwidth}{!}{%
\begin{tabular}{@{}ccccccc@{}}
\toprule
\textbf{Library} & \textbf{Primary Dependency} & \textbf{Deployment Size} & \textbf{Mathematical Scope} & \textbf{Differentiable} & \textbf{Use Case} & \textbf{Deprecated} \\ \midrule
SciPy \cite{scipy} & NumPy, C-Ext & Medium ($\sim$100MB) & General Rigid & No & General Science & No \\
SpatialMath \cite{spatialmath} & SciPy, Matplotlib & Heavy Dependencies & Robotics Superset & No & Education & No \\
PyPose \cite{pypose} & PyTorch & Massive ($>$1GB) & Diff. Manifolds & Yes & Deep Learning & No \\
tf2\_py \cite{Rosgeometry2} & ROS & Heavy ($>$2GB) & Frame Transforms & No & ROS-Based Systems & No \\
transformations.py \cite{transformationspy} & NumPy & Minimal ($\sim$MB) & Matrix Ops & No & Embedded Systems & Yes \\ 
\textbf{SE3Kit (Ours)} & \textbf{NumPy} & \textbf{Minimal ($\sim$MB)} & \textbf{$\SE$ / Lie Theory} & \textbf{No} & \textbf{Embedded Systems / Research} & \textbf{No} \\
\bottomrule
\end{tabular}%
}
\end{table}

\section{Statement of Need}
SE3Kit is an essential choice in the following work: \cite{}

\begin{itemize}
    \item \textbf{Resource-Constrained Embedded Systems:} In environments utilizing low-power embedded processors (e.g., ARM Cortex-M series or single-board computers with limited RAM), the computational overhead and dependency requirements of \texttt{PyTorch} or \texttt{Matplotlib} are prohibitive. SE3Kit implements essential kinematic operations with minimal memory allocation and low CPU utilization.
    
    \item \textbf{Cross-Platform Portability:} Unlike frameworks such as \texttt{tf2\_py}, which necessitate the Robot Operating System (ROS) build environment and specific middleware, SE3Kit is implemented in platform-independent Python. This architecture facilitates seamless integration into standalone Windows applications, web-based backends, and automated CI/CD pipelines without external compilation dependencies.
    
    \item \textbf{Hardware Calibration and Validation:} Experimental validation requires precise extrinsic calibration, including pivot calibration and hand-eye registration. SE3Kit provides verified implementations of these algorithms, enabling rigorous system characterization without the overhead of comprehensive computer vision libraries.
\end{itemize}

\section{Examples}
This work has been used in the development and experimental studies of the following research articles \cite{Maroufi2025S3DAS, Maroufi2026ComparativeAO, Maroufi2026TowardsCO, Rezayof2024OnTP, Rezayof2024QuantitativeEO, Rezayof2025ANC, Sharma2025TowardsET, Yang2024DevelopmentAQ}.

\section{Conclusion}
Although many standard tools are available in the field for 3D manipulation and calculus, there remains a critical need for a pure, lightweight, and mathematically rigorous library for 3D transformations. SE3Kit addresses this gap, decoupling fundamental geometry from the complexities of simulation engines and middleware.

\bibliographystyle{IEEEtran}
\bibliography{main}

\end{document}